\documentclass{article}

\usepackage{arxiv}

\usepackage[utf8]{inputenc} % allow utf-8 input
\usepackage[T1]{fontenc}    % use 8-bit T1 fonts
\usepackage{hyperref}       % hyperlinks
\usepackage{url}            % simple URL typesetting
\usepackage{booktabs}       % professional-quality tables
\usepackage{amsfonts}       % blackboard math symbols
\usepackage{nicefrac}       % compact symbols for 1/2, etc.
\usepackage{microtype}      % microtypography
\usepackage{lipsum}
\usepackage{graphicx}

\usepackage{amsmath}
\usepackage{multirow}
\usepackage{enumitem}
\usepackage{xcolor}
\graphicspath{ {./images/} }

\title{VFGS-Net: Frequency-Guided State-Space Learning for Topology-Preserving Retinal Vessel Segmentation}

\author{
 Ruiqi Song \\
  College of Computer Science\\
  Sichuan Normal University\\
  Chengdu, Sichuan 610068, China \\
  \texttt{20241302002@stu.sicnu.edu.cn} \\
   \And
 Lei Liu \\
  Zhejiang University \& Ant Group \\
  Hangzhou, Zhejiang, China \\
  \texttt{liulei1497@gmail.com} \\
  \And
 Ya-Nan Zhang \\
  College of Computer Science\\
  Sichuan Normal University\\
  Chengdu, Sichuan 610068, China \\
  \texttt{zyn962464@gmail.com} \\
  \And
 Chao Wang \\
  School of Engineering\\
  Sichuan Normal University\\
  Chengdu, Sichuan 610068, China \\
  \texttt{chaowangcw@hotmail.com} \\
  \And
 Xiaoning Li \\
  College of Computer Science\\
  Sichuan Normal University\\
  Chengdu, Sichuan 610068, China \\
  \texttt{lxn@sicnu.edu.cn} \\
  \And
 Nan Mu \\
  College of Computer Science\\
  Sichuan Normal University\\
  Chengdu, Sichuan 610068, China \\
  \texttt{nanmu@sicnu.edu.cn} \\
}

\begin{document}
\maketitle
\begin{abstract}
Accurate retinal vessel segmentation is a critical prerequisite for quantitative analysis of retinal images and computer-aided diagnosis of vascular diseases such as diabetic retinopathy. However, the elongated morphology, wide scale variation, and low contrast of retinal vessels pose significant challenges for existing methods, making it difficult to simultaneously preserve fine capillaries and maintain global topological continuity. To address these challenges, we propose the Vessel-aware Frequency-domain and Global Spatial modeling Network (VFGS-Net), an end-to-end segmentation framework that seamlessly integrates frequency-aware feature enhancement, dual-path convolutional representation learning, and bidirectional asymmetric spatial state-space modeling within a unified architecture. Specifically, VFGS-Net employs a dual-path feature convolution module to jointly capture fine-grained local textures and multi-scale contextual semantics. A novel vessel-aware frequency-domain channel attention mechanism is introduced to adaptively reweight spectral components, thereby enhancing vessel-relevant responses in high-level features. Furthermore, at the network bottleneck, we propose a bidirectional asymmetric Mamba2-based spatial modeling block to efficiently capture long-range spatial dependencies and strengthen the global continuity of vascular structures. Extensive experiments on four publicly available retinal vessel datasets demonstrate that VFGS-Net achieves competitive or superior performance compared to state-of-the-art methods. Notably, our model consistently improves segmentation accuracy for fine vessels, complex branching patterns, and low-contrast regions, highlighting its robustness and clinical potential.
\end{abstract}

% keywords can be removed

\section{Introduction}

% Retinal vessel-related diseases, such as diabetic retinopathy~\cite{yau2012global}, hypertensive retinopathy~\cite{wong2007eye}, and glaucoma~\cite{flammer1994vascular}, are among the leading causes of visual impairment and blindness worldwide. Early diagnosis and timely clinical intervention for these diseases rely heavily on precise analysis of retinal vascular morphology. Numerous clinical studies have demonstrated that vessel-related morphological characteristics, including vessel width, branching patterns, tortuosity, and density distribution, serve as critical biomarkers for disease screening, staging, and progression assessment~\cite{cheung2012retinal}. Consequently, accurate and robust retinal vessel segmentation constitutes a fundamental prerequisite for automated retinal image analysis and computer-aided diagnosis.

Retinal vessel-related diseases, such as diabetic retinopathy~\cite{yau2012global}, hypertensive retinopathy~\cite{wong2007eye}, and glaucoma~\cite{flammer1994vascular}, are among the leading causes of visual impairment and blindness worldwide. Early diagnosis and timely clinical intervention for these conditions critically depend on precise analysis of retinal vascular morphology. Extensive clinical studies have established that morphological characteristics of retinal vessels, including width, branching patterns, tortuosity, and density distribution, serve as key biomarkers for disease screening, staging, and monitoring progression~\cite{cheung2012retinal}. Consequently, accurate and robust retinal vessel segmentation becomes a fundamental prerequisite for automated retinal image analysis and computer-aided diagnosis.

\begin{figure}[!ht]
\centering
\includegraphics[width=0.95\columnwidth]{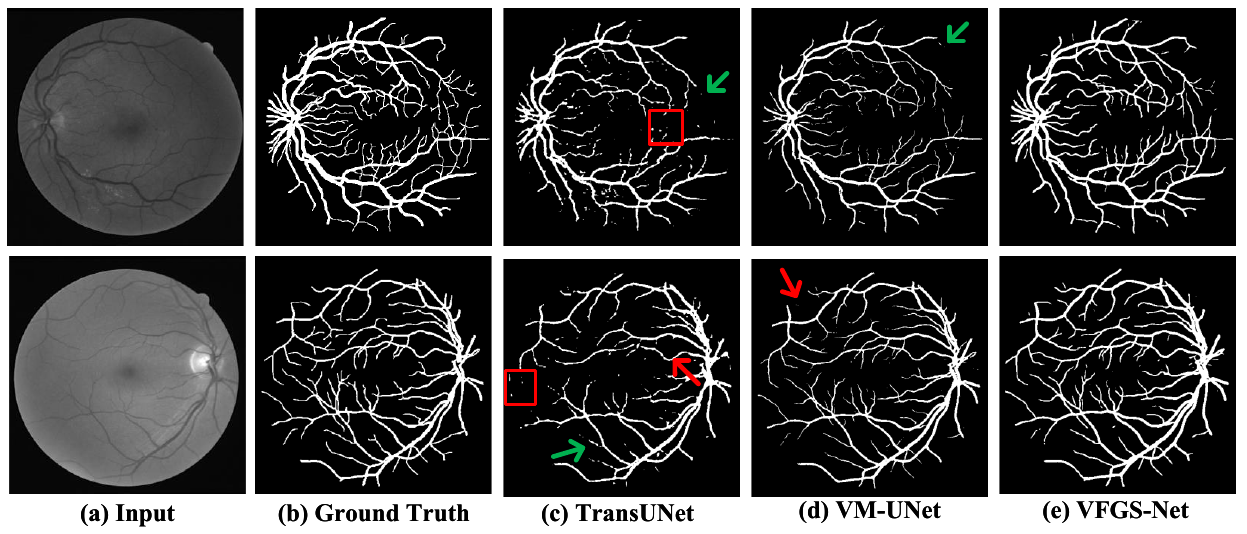} 
\vspace{-2mm} % Adjusts space between figure and caption
\caption{Representative failure cases of baseline retinal vessel segmentation models. (a) Input images; (b) Ground truth; (c-d) Results of TransUNet~\cite{chen2021transunet} and VM-UNet~\cite{ruan2024vm}, showing missed or fragmented fine vessels (highlighted); (e) Result of the proposed VFGS-Net with improved vessel continuity.}
\vspace{-4mm} % Adjusts space after caption
\label{fig:title_fig}
\end{figure} 

Despite its clinical significance, retinal vessel segmentation remains a highly challenging task. The retinal vasculature typically exhibits elongated structures, large-scale variations, and complex branching topologies~\cite{fraz2012blood}. Compounding these intrinsic difficulties, factors such as imaging noise, uneven illumination, and low vessel-to-background contrast further hinder accurate vessel delineation. Consequently, manual pixel-wise annotation is not only time-consuming and labor-intensive but also suffers from substantial inter-observer variability, severely impeding large-scale clinical deployment~\cite{verma2024systematic}. To address these challenges, deep learning-based automatic segmentation methods have garnered increasing attention. However, existing approaches still struggle to simultaneously preserve fine vessels, faithfully model intricate vascular topology, and generalize robustly across diverse datasets~\cite{qin2024review}.

Existing retinal vessel segmentation methods are mainly based on convolutional neural networks (CNNs)~\cite{lecun2002gradient}, particularly the U-Net~\cite{ronneberger2015u} and its variants, such as TransUNet~\cite{chen2021transunet} and VM-UNet~\cite{ruan2024vm}. However, these approaches remain hindered by several limitations. Firstly, the limited receptive field of standard convolutions impedes effective modeling of long-range spatial dependencies, often resulting in missed or fragmented vessel segments in morphologically complex regions (green arrows in Figure~\ref{fig:title_fig}(c-d)). Secondly, repeated downsampling and pooling operations inevitably discard fine-grained spatial details, making it difficult to preserve thin terminal vessels and capillary structures (red arrows in Figure~\ref{fig:title_fig}(c-d)). Thirdly, without strong global contextual constraints, the high visual similarity between vessels and background frequently leads to false positives and topological discontinuities, especially in low-contrast areas (red boxes in Figure~\ref{fig:title_fig}(c)).

To address the aforementioned challenges, we propose \textbf{V}essel-aware \textbf{F}requency-domain and \textbf{G}lobal \textbf{S}patial modeling Network (VFGS-Net), an end-to-end retinal vessel segmentation framework that unifies frequency-aware feature enhancement with global spatial dependency modeling in a single architecture. VFGS-Net is specifically designed to simultaneously preserve fine-grained vessel details and capture long-range structural dependencies inherent in complex vascular topologies. Specifically, we introduce a Dual-Path Feature Convolution (DFC) module that jointly models local texture cues and multi-scale contextual semantics while maintaining high-resolution feature representations, thereby enhancing the characterization of vessels across diverse scales. Additionally, we propose a Vessel-aware Frequency-domain Channel Attention (VFCA) module that adaptively amplifies vessel-relevant spectral components in high-level features, significantly improving the discriminability of thin vessels and low-contrast branches. Moreover, to effectively model the strong directionality and long-range connectivity of retinal vasculature, we integrate a bidirectional asymmetric Mamba2-based spatial modeling module at the bottleneck. This design enables efficient aggregation of global contextual information and explicitly reinforces topological continuity along vascular pathways.

The main contributions of this work are summarized as follows:
\begin{itemize}
    \item We propose VFGS-Net, a novel retinal vessel segmentation network that jointly integrates dual-path convolutional feature modeling, frequency-aware channel attention, and bidirectional asymmetric state-space modeling, providing a unified solution for accurate segmentation of complex vascular structures.
    \item We introduce a vessel-aware frequency-domain channel attention mechanism that adaptively enhances vessel-related feature responses, effectively improving the segmentation of fine vessels and low-contrast branches.
    \item We design a bidirectional asymmetric Mamba2-based spatial modeling strategy to efficiently capture long-range spatial dependencies and global vascular topology, alleviating vessel discontinuities and fragmented predictions.
    \item Extensive experiments on four public retinal vessel datasets demonstrate that the proposed method consistently outperforms state-of-the-art approaches, particularly in preserving fine vessels, modeling complex branching structures, and maintaining vascular continuity, indicating strong robustness and generalization capability.
\end{itemize}

\section{Related Work}

% Recent advances in medical image segmentation focused on enhancing spatial modeling to capture long-range dependencies and complex anatomical structures. Accordingly,

This section reviews related work on spatial modeling and retinal vessel segmentation.

\subsection{Spatial Modeling}

To overcome the inherent locality of CNNs, attention mechanisms have been widely adopted in medical image segmentation to explicitly model long-range dependencies. A representative example is TransUNet by Chen~\emph{et al.}~\cite{chen2021transunet}, which integrates Transformer modules into the U-Net encoder to enhance global contextual modeling. Building on this paradigm, numerous hybrid CNN-Transformer architectures have since emerged, ~\emph{e.g.}, Swin-Unet~\cite{cao2022swin}, ARU-Net~\cite{mu2023attention}, and FACU-Net~\cite{mu2023exploring}, which aim to combine the local representational power of convolutions with the global reasoning capability of attention. Despite their success, these attention-based models often incur high computational complexity and struggle to scale efficiently to high-resolution feature maps, limiting their applicability in fine-grained tasks like retinal vessel segmentation.

More recently, state space models (SSMs) emerged as a promising alternative for modeling long-range dependencies with linear computational complexity. Notable SSM-based approaches include S4~\cite{gu2021efficiently} and the selectively parameterized Mamba model~\cite{gu2024mamba}. In medical image segmentation, Ruan~\emph{et al.} proposed VM-UNet~\cite{ruan2024vm}, which incorporates SSMs into a U-shaped architecture to improve structural continuity in complex anatomical regions. Wang~\emph{et al.}~\cite{wang2024mamba} further introduced MambaUNet, leveraging Mamba-based modeling to efficiently capture global context while preserving local spatial fidelity. To address computational efficiency, Zheng~\emph{et al.}~\cite{wu2025ultralight} developed the UltraLight series, featuring lightweight Mamba designs tailored for large-scale applications. Zhang~\emph{et al.}~\cite{zhang2024hmt} presented HMT-UNet, which fuses Mamba with Transformer modules to strengthen global modeling; however, its capacity to retain fine local details remains limited, which is critical for resolving thin retinal vessels.

\subsection{Retinal Vessel Segmentation}

Retinal vessel segmentation is a fundamental task in computer-aided diagnosis of retinal diseases and has been extensively explored in the literature. Early and mainstream approaches are predominantly built upon U-Net variants, leveraging encoder-decoder architectures with multi-scale feature fusion to capture overall vascular structures. For example, IterNet by Li~\emph{et al.}~\cite{li2020iternet} enhances vessel continuity through iterative refinement, while CS$^2$Net by Mou~\emph{et al.}~\cite{mou2021cs2} improves fine vessel detection via cascaded supervision and contextual feature fusion. More recently, Qi~\emph{et al.} proposed DSCNet~\cite{qi2023dynamic}, which employs direction-sensitive convolutions and structure-aware modules to better model slender vessels. In parallel, Wang~\emph{et al.}~\cite{wang2025serp} introduced Serp-Mamba, integrating Mamba-based sequence modeling with direction-aware scanning to improve structural consistency in curved and crossing vessel regions.

Despite these advances, significant challenges persist, particularly in preserving fine capillaries, accurately representing complex branching topologies, and maintaining cross-scale structural coherence. Low-contrast regions and vessel crossings remain especially vulnerable to fragmentation, missed detections, and false positives.

In summary, existing methods primarily fall into two categories: (1) those that employ attention mechanisms or state space models to capture long-range dependencies, and (2) those that incorporate task-specific architectural priors to enhance vessel representation. However, attention-based approaches often suffer from high computational overhead when applied to high-resolution retinal images, while current SSM-based and structure-aware methods still fall short in jointly preserving fine-scale details, modeling intricate vascular topology, and enforcing global continuity. These limitations motivate our proposed VFGS-Net, which unifies dual-path feature convolution, frequency-domain feature modulation, and bidirectional asymmetric spatial modeling to achieve accurate, robust, and topology-preserving retinal vessel segmentation.

\section{The Proposed VFGS-Net}

\subsection{Overview}

\begin{figure*}[!t]
\centering
\includegraphics[width=\linewidth]{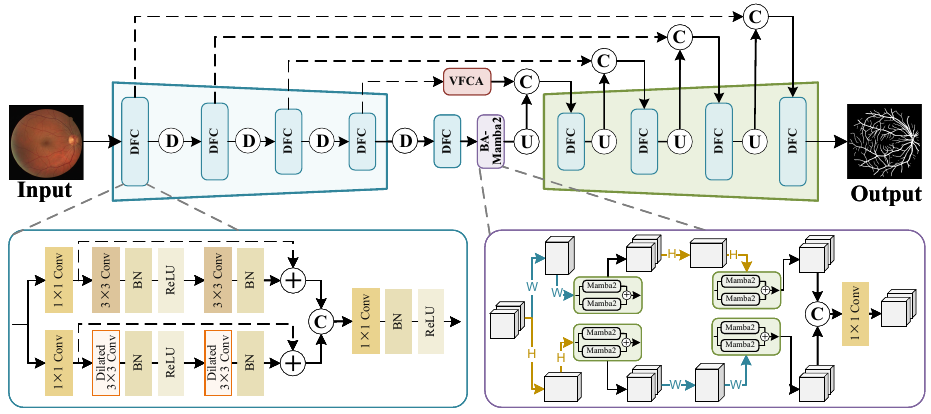} 
\vspace{-4mm} 
\caption{A schematic overview of the proposed VFGS-Net. The architecture consists of an encoder-decoder backbone with embedded modules (\emph{i.e.}, DFC, VFCA, and BA-Mamba2) designed to improve semantic representation and structural integrity.}
\vspace{-4mm} 
\label{fig:VFGS-Net}
\end{figure*} 

As illustrated in Figure~\ref{fig:VFGS-Net}, we propose VFGS-Net, an end-to-end framework for retinal vessel segmentation that aims to simultaneously preserve fine vessel details and model complex vascular topologies. The network follows an encoder-decoder paradigm and integrates three complementary modules: Dual-path Feature Convolution (DFC), Vessel-aware Frequency Channel Attention (VFCA), and Bidirectional Asymmetric Mamba2 (BA-Mamba2). These components collaboratively enable unified modeling of local vessel textures, frequency-aware vessel representations, and long-range spatial dependencies.

Specifically, the DFC module is embedded throughout both the encoder and decoder stages. Its parallel dual-path design allows the network to capture fine-grained vessel textures and broader contextual semantics simultaneously, thereby enhancing multi-scale vessel representations. To further improve sensitivity to low-contrast and thin vessels, the VFCA module adaptively emphasizes vessel-relevant responses in the frequency domain, mitigating the loss of fine details caused by repeated downsampling and convolutional smoothing. At the bottleneck, the BA-Mamba2 module performs efficient global spatial modeling via bidirectional asymmetric paths, effectively capturing long-range dependencies and reinforcing vascular connectivity and continuity. Through the coordinated interaction of these modules within a unified end-to-end framework, VFGS-Net achieves robust segmentation performance on both thin vessels and complex branching structures.

\subsection{Encoder-Decoder Structure}

VFGS-Net adopts a symmetric encoder-decoder architecture, as shown in Figure~\ref{fig:VFGS-Net}. The encoder consists of four hierarchical encoding stages, each comprising a DFC module followed by spatial downsampling. At the $i$-th encoder stage ($i \in \{1,2,3,4\}$), the input feature map is denoted as: $\mathbf{X}_i \in \mathbb{R}^{C_i \times H_i \times W_i}$.

Each encoder stage first applies the DFC module to enhance both local vessel textures and contextual semantic information:
\begin{equation}
\mathbf{E}_i = \mathrm{DFC}(\mathbf{X}_i), \quad i = 1,2,3,4.
\label{eq:encoder_dfc}
\end{equation}

Subsequently, a $2 \times 2$ max-pooling operation with stride 2 is employed to reduce the spatial resolution while preserving the channel dimension:
\begin{equation}
\mathbf{X}_{i+1} = \mathrm{MaxPool2d}(\mathbf{E}_i), \quad i = 1,2,3,4.
\label{eq:encoder_pool}
\end{equation}

Through progressive downsampling, the encoder extracts increasingly abstract and semantically rich representations while retaining essential vascular structures.

The deepest encoder feature is then forwarded to the bottleneck, where the BA-Mamba2 module is introduced to model long-range spatial dependencies:
\begin{equation}
\hat{\mathbf{E}}_5 = \mathrm{BA\text{-}Mamba2}(\mathbf{E}_5).
\label{eq:bottleneck}
\end{equation}

The decoder mirrors the encoder with four symmetric decoding stages, aiming to progressively restore spatial resolution and reconstruct fine vascular structures. At the $i$-th decoding stage ($i \in \{1,2,3,4\}$), the feature map from the previous stage is first upsampled via bilinear interpolation and then concatenated with the corresponding encoder feature through a skip connection:
\begin{equation}
\mathbf{Y}_i = \mathrm{Concat}\!\left(\mathrm{Interp}(\mathbf{D}_{i+1}), \mathbf{E}_i \right).
\label{eq:decoder_concat}
\end{equation}

The fused features are subsequently refined using a DFC module:
\begin{equation}
\mathbf{D}_i = \mathrm{DFC}(\mathbf{Y}_i).
\label{eq:decoder_dfc}
\end{equation}

This decoding strategy effectively integrates shallow texture details from the encoder with deep semantic information from the bottleneck, thereby preserving vessel continuity and topological consistency during spatial reconstruction.

Notably, at the deepest skip connection, the encoder feature is further enhanced by the proposed VFCA module prior to feature fusion. This operation adaptively emphasizes vessel-related frequency components, improving sensitivity to low-contrast regions and fine vascular branches.

\subsection{DFC Module}

To effectively capture both fine-grained vessel textures and broader contextual structures, VFGS-Net employs the DFC module as the fundamental feature extraction unit in both the encoder and decoder. As illustrated in Figure~\ref{fig:VFGS-Net}, DFC adopts a parallel dual-path architecture consisting of a standard convolution path and a dilated convolution path, enabling complementary modeling of local and cross-scale features within a unified residual framework.

Given an input feature map $\mathbf{X} \in \mathbb{R}^{C \times H \times W}$, the input is first projected into an intermediate feature space via a shared $1 \times 1$ convolution, which also serves as the residual shortcut for both paths:
\begin{equation}
\mathbf{X}_p = \mathrm{Conv}_{1 \times 1}(\mathbf{X}).
\label{eq:dfc_proj}
\end{equation}

In the standard convolution path, local texture features are extracted using two successive $3 \times 3$ convolutions, each followed by batch normalization (BN), with a ReLU activation applied between the two layers. A residual connection is incorporated to preserve low-level vessel information and stabilize optimization:
\begin{equation}
\mathbf{R} = \mathrm{BN}\!\left(\mathrm{Conv}_{3 \times 3}\!\left(
\mathrm{ReLU}\!\left(\mathrm{BN}\!\left(\mathrm{Conv}_{3 \times 3}(\mathbf{X}_p)\right)\right)
\right)\right) + \mathbf{X}_p .
\label{eq:dfc_std}
\end{equation}

In parallel, the dilated convolution path follows an identical residual structure but replaces standard convolutions with dilated $3 \times 3$ convolutions using a dilation rate of $d=2$, thereby enlarging the receptive field and capturing richer contextual information:
\begin{equation}
\mathbf{D} = \mathrm{BN}\!\left(\mathrm{Conv}_{3 \times 3}^{(d=2)}\!\left(
\mathrm{ReLU}\!\left(\mathrm{BN}\!\left(\mathrm{Conv}_{3 \times 3}^{(d=2)}(\mathbf{X}_p)\right)\right)
\right)\right) + \mathbf{X}_p .
\label{eq:dfc_dilated}
\end{equation}

The outputs from the standard and dilated paths are concatenated along the channel dimension and fused through a $1 \times 1$ convolution followed by batch normalization and ReLU activation:
\begin{equation}
\mathbf{Y} = \mathrm{ReLU}\!\left(\mathrm{BN}\!\left(\mathrm{Conv}_{1 \times 1}\!\left(\mathrm{Concat}(\mathbf{R}, \mathbf{D})\right)\right)\right).
\label{eq:dfc_fusion}
\end{equation}

By jointly modeling fine local textures and enlarged contextual dependencies through complementary convolution paths, the DFC module effectively enhances the representation of thin vessels, complex branching patterns, and vascular continuity, providing robust feature foundations for subsequent encoding and decoding processes.

\subsection{VFCA Module}

\begin{figure}[ht]
\centering
\includegraphics[width=0.9\linewidth]{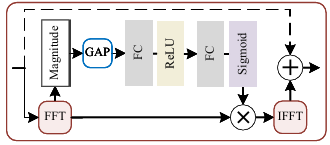}
\vspace{-2mm} 
\caption{Architecture of the proposed VFCA module, which performs channel-wise attention in the frequency domain via FFT-based global magnitude pooling and adaptive frequency reweighting.}
\vspace{-2mm} 
\label{fig:vfca}
\end{figure}

To further enhance the decoder’s ability to represent fine vascular structures, we propose the VFCA module and embed it into the skip connection from the fourth encoder stage. Unlike conventional channel attention mechanisms that operate purely in the spatial domain, VFCA explicitly models channel-wise importance in the frequency domain, enabling adaptive enhancement of vessel-related frequency components while suppressing background noise and irrelevant textures. The architecture of the VFCA module is illustrated in Figure~\ref{fig:vfca}.

Let $\mathbf{R}_4 \in \mathbb{R}^{B \times C \times H \times W}$ denote the feature map extracted from the fourth encoder stage, where $B$, $C$, $H$, and $W$ represent the batch size, number of channels, height, and width, respectively. VFCA first transforms the spatial features into the frequency domain using a two-dimensional fast Fourier transform (FFT, denoted as $\mathcal{F}$), applied independently to each channel:
\begin{equation}
\mathbf{F}_4 = \mathcal{F}(\mathbf{R}_4).
\label{eq:fft}
\end{equation}

Next, VFCA extracts a global frequency descriptor by computing the magnitude spectrum of $\mathbf{F}_4$ and performing global average pooling (GAP) over the spatial frequency dimensions:
\begin{equation}
f_c = \frac{1}{HW} \sum_{h=1}^{H} \sum_{w=1}^{W} \left| \mathbf{F}_4(c,h,w) \right|, 
\quad c = 1,2,\ldots,C .
\label{eq:freq_pool}
\end{equation}

This operation summarizes the overall frequency response of each channel, capturing vessel-related texture and structural information that is less sensitive to spatial variations.

The resulting frequency descriptors are then fed into a lightweight two-layer fully connected network with a bottleneck structure to generate channel-wise attention weights:
\begin{equation}
\boldsymbol{\alpha} = \delta\!\left( \mathbf{W}_2 \, \sigma\!\left( \mathbf{W}_1 \mathbf{f} \right) \right),
\label{eq:attention}
\end{equation}
where $\mathbf{f} = [f_1, f_2, \ldots, f_C]^{\top}$, $\mathbf{W}_1$ and $\mathbf{W}_2$ are learnable linear transformations, $\sigma(\cdot)$ denotes the ReLU activation, and $\delta(\cdot)$ denotes the Sigmoid function. The attention vector $\boldsymbol{\alpha} \in [0,1]^C$ reflects the relative importance of each channel in the frequency domain.

Subsequently, the channel attention weights are applied to the frequency-domain features via channel-wise multiplication:
\begin{equation}
\widetilde{\mathbf{F}}_4(c,h,w) = \alpha_c \cdot \mathbf{F}_4(c,h,w),
\label{eq:freq_reweight}
\end{equation}
followed by an inverse FFT to project the enhanced features back to the spatial domain:
\begin{equation}
\widehat{\mathbf{R}}_4 = \mathcal{F}^{-1}(\widetilde{\mathbf{F}}_4),
\label{eq:ifft}
\end{equation}
where $\mathcal{F}^{-1}(\cdot)$ denotes the inverse FFT. The refined feature map $\widehat{\mathbf{R}}_4$ is forwarded to the decoder via skip connections, providing frequency-aware and vessel-enhanced representations. By emphasizing vessel-sensitive channels under global frequency guidance, the VFCA module improves the segmentation of thin vessels and complex branches, particularly in low-contrast regions.

\begin{figure}[ht]
\centering
\includegraphics[width=0.95\linewidth]{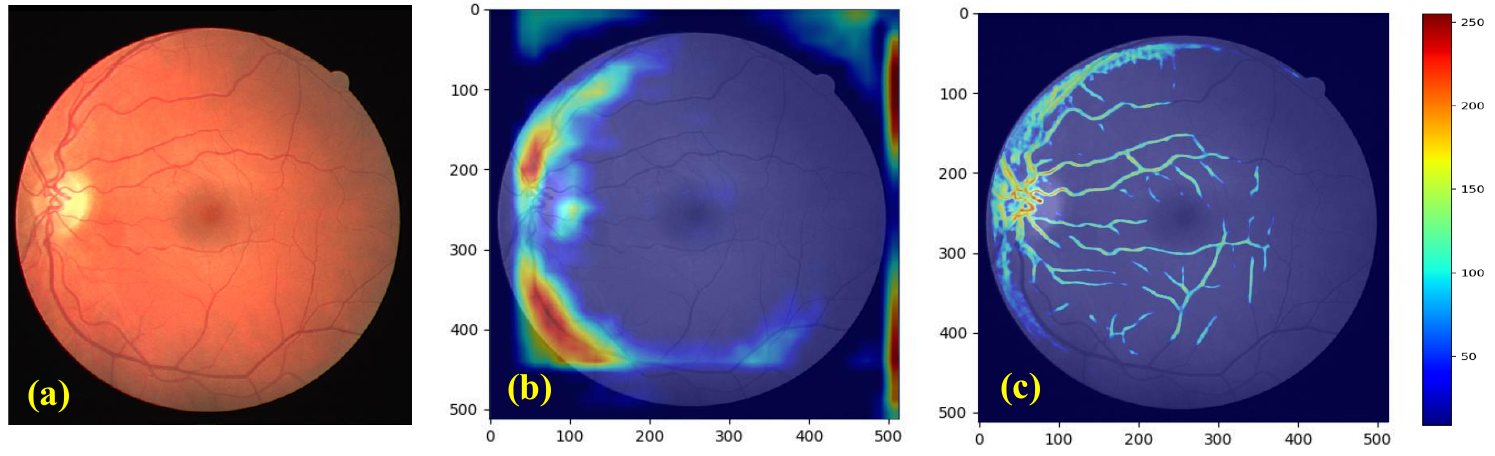}
\vspace{-2mm} 
\caption{Comparative Grad-CAM~\cite{selvaraju2017grad} visualizations  of standard skip connections and the proposed VFCA module. (a) Input image. (b-c) Heatmaps generated by standard skip connections and VFCA module, respectively. Red indicates strong activation and blue indicates weak activation.}
\vspace{-2mm} 
\label{fig:gradcam_vfca}
\end{figure}

A visual comparison between standard skip connections and the proposed VFCA module is shown in Figure~\ref{fig:gradcam_vfca}. Standard skip connections exhibit local discontinuities in major vessels, missing branch structures, and noticeable background interference. In contrast, VFCA modulates skip-connection features in the frequency domain, improving vessel continuity, branch preservation, and background suppression. Grad-CAM responses further indicate that low-frequency components capture global vascular topology, while high-frequency components emphasize thin vessels and branch details. Through frequency-aware channel reweighting, VFCA enables more effective feature integration during upsampling, yielding clearer and more continuous vascular representations.

\begin{figure}[b]
\centering
\includegraphics[width=1.0\linewidth]{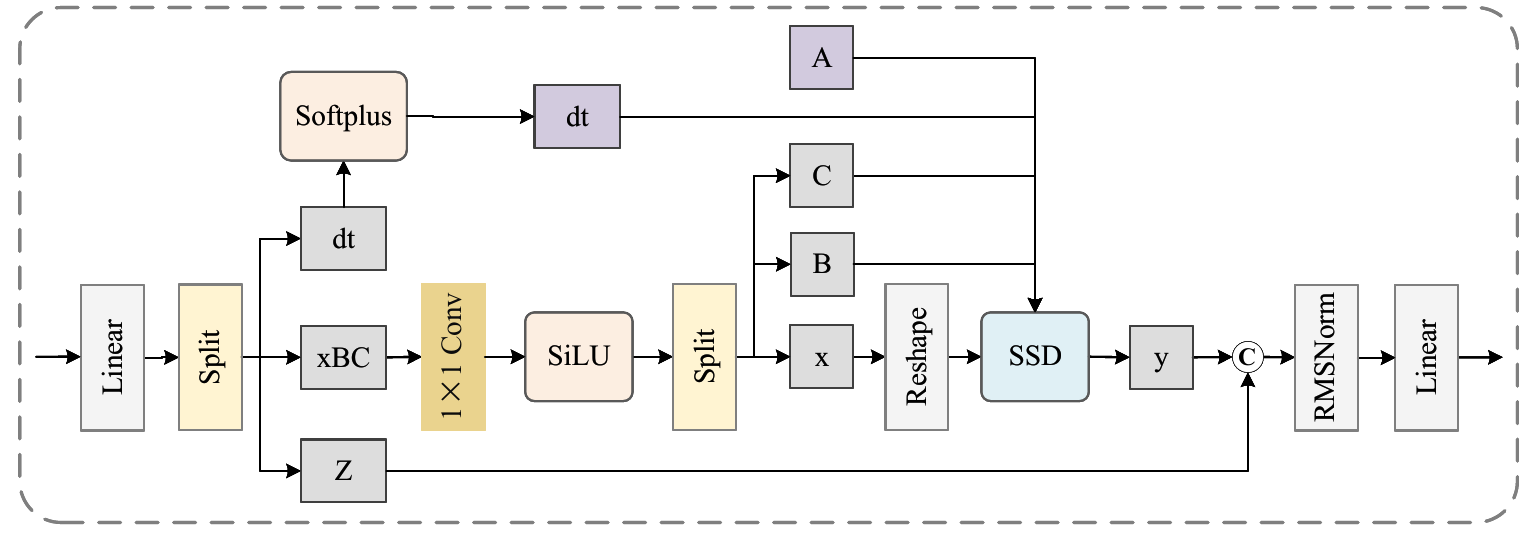}
\vspace{-2mm} 
\caption{Architecture of the Mamba2 module.}
\vspace{-2mm} 
\label{fig:mamba2}
\end{figure}

\subsection{BA-Mamba2 Module}

To effectively model global spatial dependencies in complex vascular topologies, we introduce the Bidirectional Asymmetric Mamba2 (BA-Mamba2) module at the bottleneck of VFGS-Net. Unlike self-attention mechanisms with quadratic computational complexity, BA-Mamba2 reformulates two-dimensional spatial modeling as a sequence modeling problem based on state-space representations, enabling efficient long-range dependency aggregation while preserving vessel continuity and fine structural details.

Given an input feature map $\mathbf{X} \in \mathbb{R}^{B \times C \times H \times W}$, BA-Mamba2 adopts an alternating axial decomposition strategy to factorize 2D spatial dependencies into two asymmetric one-dimensional scanning paths. Specifically, the feature map is reshaped into width-wise sequences by treating each spatial row independently, $\mathbf{X}_w \in \mathbb{R}^{(B \cdot H) \times C \times W},$
and into height-wise sequences by treating each spatial column independently, $ \mathbf{X}_h \in \mathbb{R}^{(B \cdot W) \times C \times H}.$ This axial decomposition enables effective modeling of long-range dependencies along both horizontal and vertical directions while avoiding the prohibitive computational cost of full 2D attention.

Along each axial sequence, BA-Mamba2 employs bidirectional Mamba2 modeling to enhance sensitivity to vessel continuity and symmetric structures. For a generic sequence $\mathbf{S}$, bidirectional modeling is formulated as
\begin{equation}
\mathcal{M}_{\mathrm{bi}}(\mathbf{S}) = \mathcal{M}(\mathbf{S}) + \mathcal{M}(\mathrm{Flip}(\mathbf{S})),
\label{eq:bi_mamba}
\end{equation}
where $\mathcal{M}(\cdot)$ denotes the Mamba2 state-space model and $\mathrm{Flip}(\cdot)$ reverses the sequence order along the scanning dimension. The detailed architecture of Mamba2~\cite{dao2024transformers} is illustrated in Figure~\ref{fig:mamba2}. This bidirectional operator is applied to both axial sequences, producing $ \mathbf{Y}_{wh} = \mathcal{M}_{\mathrm{bi}}(\mathbf{X}_w)$, and $\mathbf{Y}_{hw} = \mathcal{M}_{\mathrm{bi}}(\mathbf{X}_h). 
$

The outputs from the two asymmetric paths are then reshaped back to the original spatial resolution, concatenated along the channel dimension, and fused via a $1 \times 1$ convolution:
\begin{equation}
\mathbf{Y}_{\mathrm{out}} =
\mathrm{Conv}_{1 \times 1}\!\left(
\mathrm{Concat}(\mathbf{Y}_{wh}, \mathbf{Y}_{hw})
\right).
\label{eq:ba_mamba_fusion}
\end{equation}

The resulting feature map encodes rich multi-directional global dependencies while preserving high-resolution structural information, providing a semantically enhanced representation for subsequent decoding.

Through this bidirectional asymmetric spatial modeling strategy, BA-Mamba2 efficiently integrates global context from multiple directions within a lightweight framework. It complements the local and frequency-aware representations learned by preceding modules, leading to improved modeling of long-range vascular connectivity and complex branching patterns.

\begin{figure}[hb]
\centering
\includegraphics[width=0.95\linewidth]{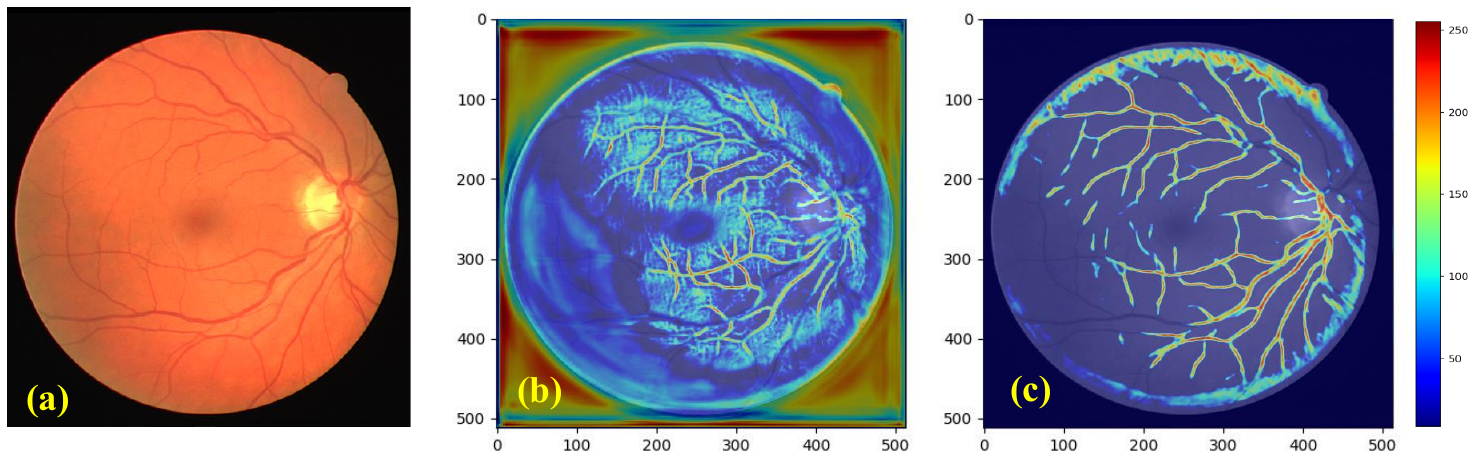}
\vspace{-2mm} 
\caption{Comparative Grad-CAM visualizations for different sequence modeling strategies. (a) Original retinal image. (b) Grad-CAM heatmaps generated by conventional Mamba2 and the proposed BA-Mamba2, respectively.}
\vspace{-2mm} 
\label{fig:gradcam_mamba}
\end{figure}

As shown in Figure~\ref{fig:gradcam_mamba}, BA-Mamba2 exhibits broader and more coherent activation regions spanning long-range vascular structures than the conventional Mamba2 module. In particular, BA-Mamba2 highlights continuous vessel trajectories and spatially distant yet correlated regions, whereas conventional Mamba2 produces more localized and fragmented activations. Moreover, bidirectional asymmetric scanning preserves precise local responses while strengthening cross-directional dependencies, resulting in clearer and more complete activations along thin vessels and branching structures. These observations demonstrate the effectiveness of BA-Mamba2 in jointly capturing global context and fine-grained vascular details.

\subsection{Loss Function}

In retinal vessel segmentation, vessel pixels typically occupy only a small portion of the image, leading to severe class imbalance between foreground and background. To address this issue, we employ a joint optimization strategy that combines weighted binary cross-entropy (BCE) loss and Dice loss, enforcing both pixel-wise classification accuracy and global vascular structural consistency. The overall segmentation loss is defined as
\begin{equation}
\mathcal{L}_{\mathrm{seg}} = \mathcal{L}_{\mathrm{BCE}} + \mathcal{L}_{\mathrm{Dice}} .
\label{eq:loss_total}
\end{equation}

The weighted BCE loss is formulated as
\begin{equation}
\mathcal{L}_{\mathrm{BCE}} =
- \frac{1}{N} \sum_{i=1}^{N}
\left[
p \, y_i \log(\hat{y}_i) +
(1 - y_i)\log(1 - \hat{y}_i)
\right],
\label{eq:loss_bce}
\end{equation}
where $N$ denotes the total number of pixels, $y_i \in \{0,1\}$ is the ground-truth label of the $i$-th pixel, $\hat{y}_i = \delta(P_i)$ is the Sigmoid-activated prediction, and $p$ is a foreground weighting factor introduced to alleviate class imbalance. The Dice loss is defined as
\begin{equation}
\mathcal{L}_{\mathrm{Dice}} =
1 -
\frac{2 \sum_{i=1}^{N} \hat{y}_i y_i + \epsilon}
{\sum_{i=1}^{N} \hat{y}_i + \sum_{i=1}^{N} y_i + \epsilon},
\label{eq:loss_dice}
\end{equation}
where $\epsilon = 10^{-7}$ is a small constant for numerical stability.

By jointly optimizing weighted BCE and Dice loss, the proposed loss function effectively balances pixel-level discrimination and global shape preservation. This design promotes accurate delineation of fine vessels and complex branching structures while maintaining vascular continuity, which is essential for reliable retinal vessel segmentation.

\section{Experiments}

This section presents the experimental setup, quantitative performance evaluation, and ablation studies conducted to validate the effectiveness and robustness of the proposed method.

\subsection{Experimental Setup}

\textbf{Training Details:}
The proposed method is implemented in PyTorch and trained on two NVIDIA RTX~3090 GPUs with a total memory of 48~GB. Model optimization is performed using the Adam optimizer~\cite{kingma2014adam} with an initial learning rate of $1\times10^{-4}$ and momentum parameters $\beta_1=0.9$ and $\beta_2=0.999$. A StepLR scheduler is adopted to halve the learning rate every 10 epochs, ensuring stable convergence. All models are trained for 300 epochs without early stopping, and intermediate checkpoints are saved periodically. To enhance generalization, online data augmentation is applied during training, including random horizontal and vertical flipping, small-angle rotations ($\pm7^{\circ}$) to preserve fine vessel continuity, and random Gamma correction to simulate illumination variations. In addition, contrast-limited adaptive histogram equalization (CLAHE)~\cite{shaout2023novel} is applied with low probability to enhance local vessel contrast while avoiding excessive noise amplification. All images are resized to $512\times512$ after augmentation. Following common practice in retinal vessel segmentation, only the green channel is used as network input due to its higher vessel-to-background contrast~\cite{fraz2012ensemble}. Corresponding annotation masks are resized using nearest-neighbor interpolation and binarized to ensure stable and consistent supervision. All experiments are conducted using the predefined training/validation/testing splits of each public dataset to ensure fair comparison and reproducibility.

\textbf{Image Datasets:}
To comprehensively evaluate segmentation accuracy and generalization ability, experiments are conducted on four widely used public retinal vessel datasets:
\begin{itemize}
    \item \textbf{DRIVE}~\cite{staal2004ridge}: consists of 40 color fundus images ($565 \times 584$ pixels) with a $45^{\circ}$ field of view, officially split into 20 training and 20 testing images.
    \item \textbf{HRF}~\cite{budai2013robust}: contains 45 high-resolution fundus images ($3504 \times 2336$ pixels) from healthy subjects and patients with diabetic retinopathy and glaucoma, posing significant challenges for multi-scale vessel modeling.
    \item \textbf{CHASE\_DB1}~\cite{fraz2012ensemble}: includes 28 pediatric fundus images (approximately $999 \times 960$ pixels) characterized by uneven illumination and low contrast, making micro-vessel segmentation particularly difficult.
    \item \textbf{STARE}~\cite{hoover2000locating}: comprises 20 fundus images ($700 \times 605$ pixels) with diverse pathological conditions, commonly used to assess robustness under complex vessel topology.
\end{itemize}

\textbf{Evaluation Metrics:}
Segmentation performance is quantitatively evaluated using five pixel-level metrics: Dice Similarity Coefficient (Dice), Sensitivity (SE), Specificity (SP), 95\% Hausdorff Distance (HD95), and Average Symmetric Surface Distance (ASSD). These metrics jointly assess both region overlap and boundary accuracy between predictions and ground-truth annotations. All results are computed on the test sets and averaged across images. Detailed metric definitions follow~\cite{mu2023attention}.

\subsection{Segmentation Performance Evaluation}

We compare the proposed VFGS-Net with six representative state-of-the-art segmentation models, including U-Net~\cite{ronneberger2015u}, DSCNet~\cite{qi2023dynamic}, HMT-UNet~\cite{zhang2024hmt}, UltraLight~\cite{wu2025ultralight}, VM-UNet~\cite{ruan2024vm}, and Serp-Mamba~\cite{wang2025serp}, using the five evaluation metrics introduced in Section~4.1.

\begin{table}[ht]
\centering
\vspace{-2mm} 
\caption{Quantitative comparison of different segmentation models on four datasets ($\uparrow$ indicates higher is better; $\downarrow$ indicates lower is better).}
\vspace{-2mm} 
%\scriptsize %footnotesize
%\setlength{\tabcolsep}{1.0pt} 
\label{tab:quantitative}
\resizebox{\columnwidth}{!}{
\begin{tabular}{c|c|c|c|c|c|c}
\hline
Dataset & Model & Dice (\%) $\uparrow$ & SE (\%) $\uparrow$ & SP (\%) $\uparrow$ & HD95 (pixel) $\downarrow$ & ASSD (pixel) $\downarrow$ \\
\hline
\multirow{7}{*}{DRIVE}
& U-Net        & 81.06$\pm$0.42 & 79.06$\pm$0.38 & 97.81$\pm$0.11 & 3.30$\pm$0.18 & 0.61$\pm$0.03 \\
& DSCNet       & 81.20$\pm$0.35 & 80.02$\pm$0.44 & \textcolor{red}{98.30$\pm$0.12} & 3.66$\pm$0.22 & 0.75$\pm$0.04 \\
& HMT-UNet     & 80.71$\pm$0.48 & 80.08$\pm$0.52 & 98.24$\pm$0.14 & 6.28$\pm$0.45 & 1.46$\pm$0.08 \\
& UltraLight   & 78.57$\pm$0.50 & 79.25$\pm$0.47 & 97.87$\pm$0.13 & 9.64$\pm$0.62 & 2.60$\pm$0.09 \\
& VM-UNet      & 80.96$\pm$0.39 & 82.19$\pm$0.43 & 98.00$\pm$0.15 & 8.99$\pm$0.37 & 1.24$\pm$0.07 \\
& Serp-Mamba   & 82.00$\pm$0.41 & 80.39$\pm$0.36 & 97.89$\pm$0.12 & 7.75$\pm$0.35 & 1.08$\pm$0.06 \\
& \textbf{VFGS-Net}& \textcolor{red}{83.23$\pm$0.33} & \textcolor{red}{82.37$\pm$0.39} & 97.87$\pm$0.11 & \textcolor{red}{2.41$\pm$0.19} & \textcolor{red}{0.55$\pm$0.03} \\
\hline
\multirow{7}{*}{HRF}
& U-Net        & 80.79$\pm$0.41 & 77.13$\pm$0.38 & 97.05$\pm$0.12 & 4.89$\pm$0.21 & 0.72$\pm$0.04 \\
& DSCNet       & 82.78$\pm$0.36 & 80.46$\pm$0.40 & 97.00$\pm$0.11 & 3.59$\pm$0.19 & 0.55$\pm$0.03 \\
& HMT-UNet     & 79.08$\pm$0.44 & 77.69$\pm$0.41 & \textcolor{red}{98.47$\pm$0.14} & 6.58$\pm$0.38 & 0.91$\pm$0.05 \\
& UltraLight   & 78.04$\pm$0.48 & 75.11$\pm$0.42 & 97.94$\pm$0.13 & 9.45$\pm$0.70 & 2.66$\pm$0.10 \\
& VM-UNet      & 79.71$\pm$0.39 & 81.22$\pm$0.41 & 98.23$\pm$0.12 & 5.97$\pm$0.26 & 0.71$\pm$0.04 \\
& Serp-Mamba   & 83.32$\pm$0.37 & 84.17$\pm$0.45 & 97.17$\pm$0.18 & 6.50$\pm$0.31 & 1.52$\pm$0.06 \\
& \textbf{VFGS-Net}& \textcolor{red}{85.60$\pm$0.33} & \textcolor{red}{84.35$\pm$0.38} & 97.26$\pm$0.11 & \textcolor{red}{2.21$\pm$0.16} & \textcolor{red}{0.43$\pm$0.03} \\
\hline
\multirow{7}{*}{CHASE\_DB1}
& U-Net        & 79.05$\pm$0.43 & 78.94$\pm$0.39 & 96.61$\pm$0.14 & 4.58$\pm$0.22 & 0.92$\pm$0.05 \\
& DSCNet       & 80.53$\pm$0.38 & 80.12$\pm$0.41 & 96.73$\pm$0.13 & 3.96$\pm$0.18 & 0.85$\pm$0.04 \\
& HMT-UNet     & 78.78$\pm$0.46 & 77.47$\pm$0.42 & \textcolor{red}{98.59$\pm$0.12} & 8.35$\pm$0.36 & 1.15$\pm$0.06 \\
& UltraLight   & 77.14$\pm$0.51 & 76.74$\pm$0.47 & 97.86$\pm$0.15 & 9.25$\pm$0.81 & 3.07$\pm$0.12 \\
& VM-UNet      & 78.52$\pm$0.40 & 79.05$\pm$0.44 & 98.36$\pm$0.13 & 7.38$\pm$0.29 & 1.16$\pm$0.05 \\
& Serp-Mamba   & 80.49$\pm$0.37 & 79.56$\pm$0.48 & 97.58$\pm$0.19 & 9.48$\pm$0.41 & 2.20$\pm$0.09 \\
& \textbf{VFGS-Net}& \textcolor{red}{81.43$\pm$0.35} & \textcolor{red}{80.69$\pm$0.40} & 97.14$\pm$0.12 & \textcolor{red}{3.60$\pm$0.17} & \textcolor{red}{0.73$\pm$0.04} \\
\hline
\multirow{7}{*}{STARE}
& U-Net        & 78.82$\pm$0.44 & 72.27$\pm$0.51 & 98.74$\pm$0.12 & 9.73$\pm$0.36 & 1.32$\pm$0.06 \\
& DSCNet       & 80.20$\pm$0.39 & 77.21$\pm$0.46 & 98.38$\pm$0.11 & 7.11$\pm$0.29 & 1.10$\pm$0.05 \\
& HMT-UNet     & 76.28$\pm$0.48 & 71.57$\pm$0.49 & \textcolor{red}{98.78$\pm$0.10} & 11.27$\pm$0.41 & 1.62$\pm$0.07 \\
& UltraLight   & 70.27$\pm$0.57 & 75.14$\pm$0.54 & 97.72$\pm$0.16 & 14.24$\pm$1.12 & 3.07$\pm$0.21 \\
& VM-UNet      & 74.59$\pm$0.45 & 72.76$\pm$0.47 & 98.31$\pm$0.13 & 13.33$\pm$0.58 & 2.05$\pm$0.08 \\
& Serp-Mamba   & 80.53$\pm$0.37 & 81.40$\pm$0.43 & 97.60$\pm$0.18 & 11.75$\pm$0.52 & 2.09$\pm$0.09 \\
& \textbf{VFGS-Net}& \textcolor{red}{83.21$\pm$0.36} & \textcolor{red}{82.42$\pm$0.40} & 98.39$\pm$0.14 & \textcolor{red}{3.44$\pm$0.26} & \textcolor{red}{0.76$\pm$0.07} \\
\hline
\end{tabular}
}
\end{table}

\textbf{Quantitative Evaluation:}
The quantitative results on four public retinal vessel datasets, namely DRIVE, HRF, CHASE\_DB1, and STARE, are summarized in Table~\ref{tab:quantitative}. Overall, VFGS-Net consistently achieves the highest Dice scores and the lowest boundary-related errors (HD95 and ASSD) across all datasets, demonstrating superior capability in preserving fine vessels and maintaining global vascular topology. On the DRIVE dataset, VFGS-Net attains a Dice score of 83.23\% with an HD95 of 2.41 pixels, outperforming Serp-Mamba (82.00\%, 7.75 pixels) and U-Net (81.06\%, 3.30 pixels). The proposed method achieves a favorable balance between sensitivity and specificity, indicating improved detection of thin vessels while effectively suppressing false positives. On the high-resolution HRF dataset, VFGS-Net further demonstrates its advantage by achieving 85.60\% Dice and 2.21 pixels HD95, substantially outperforming Serp-Mamba (83.32\%, 6.50 pixels) and VM-UNet (79.71\%, 5.97 pixels). These results highlight its effectiveness in handling high-resolution images with complex vascular patterns and pathological variations. For CHASE\_DB1 dataset, which features uneven illumination and limited training samples, VFGS-Net again delivers the best performance, achieving 81.43\% Dice and 3.60 pixels HD95. Competing methods exhibit noticeably higher boundary errors, indicating inferior robustness under challenging imaging conditions. The STARE dataset poses additional challenges due to complex vessel topology and pathological interference. While several baseline methods suffer from pronounced performance degradation, VFGS-Net maintains a high Dice score of 83.21\% and a low HD95 of 3.44 pixels, demonstrating strong robustness and generalization capability.

Overall, VFGS-Net exhibits the most balanced and stable performance across varying datasets, image qualities, and resolutions, consistently excelling in fine vessel preservation, boundary accuracy, and long-range structural continuity. These results validate the effectiveness of integrating frequency-aware feature enhancement and global spatial dependency modeling.

\begin{figure}[ht]
\centering
\includegraphics[width=1.00\linewidth]{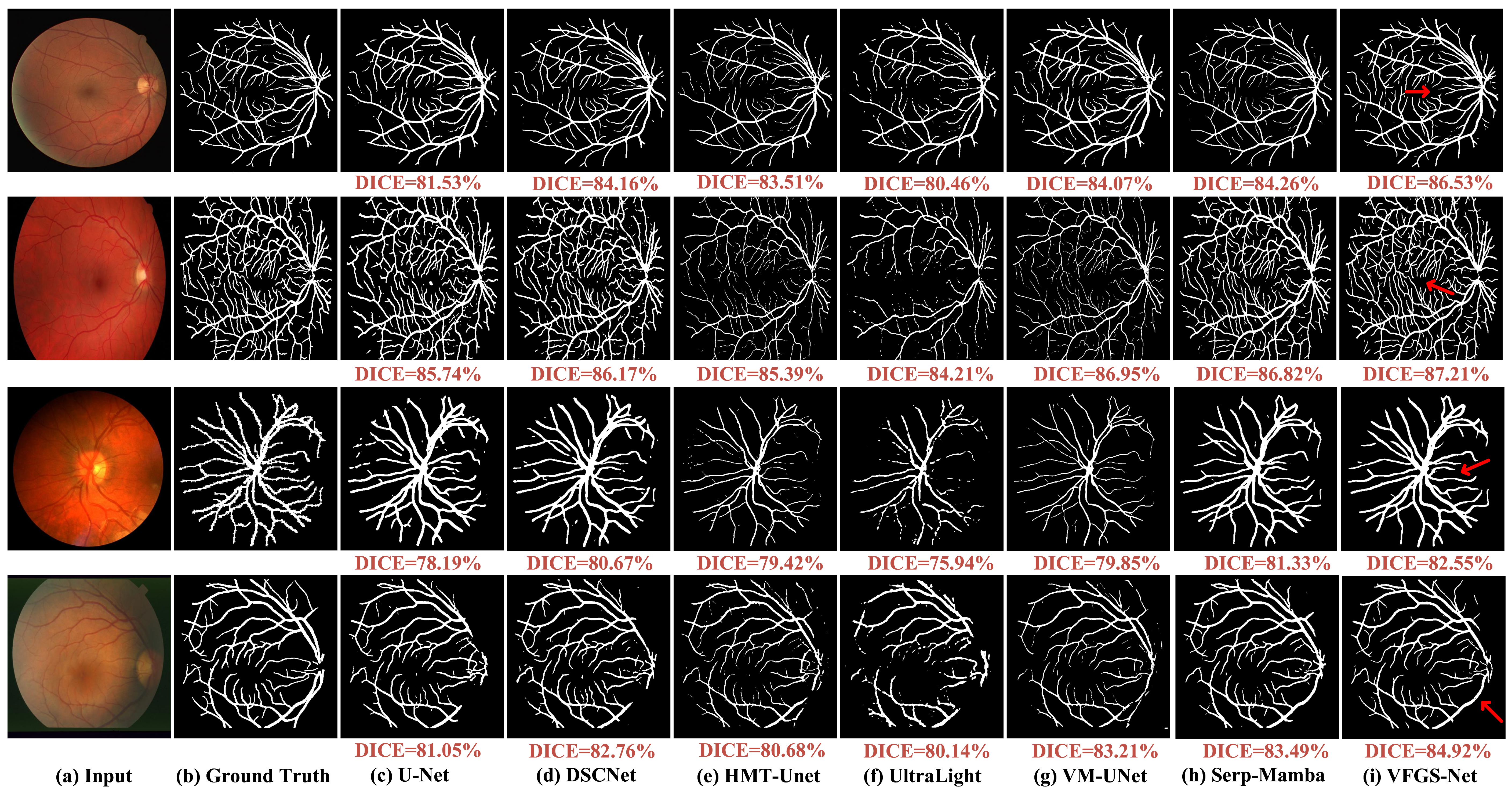}
\vspace{-2mm} 
\caption{Qualitative comparison of retinal vessel segmentation results. (a) Input images; (b) Ground truth; (c-i) Results of different methods. Dice scores are reported for reference.}
\vspace{-2mm} 
\label{fig:qualitative}
\end{figure}

\textbf{Qualitative Evaluation:}
Figure~\ref{fig:qualitative} presents representative qualitative comparisons of retinal vessel segmentation results across the four datasets. As shown, VFGS-Net consistently demonstrates superior capability in preserving vessel continuity, bifurcation structures, and fine distal branches, particularly in low-contrast regions and complex branching areas (highlighted by red arrows). In contrast, baseline methods exhibit distinct limitations. U-Net and DSCNet tend to produce fragmented vessels or blurred boundaries in low-contrast regions, leading to discontinuities along thin branches. HMT-UNet and UltraLight frequently miss fine and distal vessels, resulting in incomplete vascular trees. VM-UNet captures major vessels but introduces minor artifacts around complex crossings. Serp-Mamba preserves overall structural consistency but struggles to accurately delineate thin peripheral vessels and terminal branches. Overall, VFGS-Net generates smoother vessel contours and more coherent vascular topology, effectively reducing breakages and false negatives in challenging regions. These qualitative observations are highly consistent with the quantitative results reported in Table~\ref{tab:quantitative}, further validating the effectiveness of our method in fine vessel preservation and robust topological modeling.

\subsection{Ablation Studies}

To evaluate the contribution of each component in VFGS-Net, we design eight ablation variants based on a U-Net backbone: baseline U-Net (U); U-Net with DFC (U+D); with BA-Mamba2 (U+B); with VFCA (U+V); with DFC and BA-Mamba2 (U+D+B); with DFC and VFCA (U+D+V); with BA-Mamba2 and VFCA (U+B+V); and the full VFGS-Net (U+D+B+V).

\begin{table}[ht]
\centering
\vspace{-2mm} 
\caption{Quantitative results of different ablation variants.}
\vspace{-2mm} 
\label{tab:ablation}
\resizebox{\columnwidth}{!}{
\begin{tabular}{c|c|c|c|c|c|c}
\hline
Dataset & Model & Dice (\%)$\uparrow$ & Sensitivity (\%)$\uparrow$ & Specificity (\%)$\uparrow$ & HD95 (pixel)$\downarrow$ & ASSD (pixel)$\downarrow$ \\
\hline
\multirow{8}{*}{DRIVE}
& U & 81.06$\pm$0.42 & 79.06$\pm$0.48 & 97.81$\pm$0.14 & 3.30$\pm$0.18 & 0.61$\pm$0.04 \\
& U+D & 82.40$\pm$0.39 & 80.58$\pm$0.44 & 97.95$\pm$0.13 & 2.88$\pm$0.16 & 0.57$\pm$0.03 \\
& U+B & 81.34$\pm$0.41 & 78.18$\pm$0.46 & 98.09$\pm$0.12 & 4.27$\pm$0.21 & 0.71$\pm$0.05 \\
& U+V & 81.37$\pm$0.40 & 78.23$\pm$0.45 & 98.09$\pm$0.12 & 4.10$\pm$0.20 & 0.68$\pm$0.05 \\
& U+D+B & 81.55$\pm$0.38 & 79.55$\pm$0.43 & 97.87$\pm$0.14 & 3.25$\pm$0.17 & 0.62$\pm$0.04 \\
& U+D+V & 82.10$\pm$0.36 & 78.83$\pm$0.42 & \textcolor{red}{98.20$\pm$0.11} & 3.86$\pm$0.19 & 0.68$\pm$0.05 \\
& U+B+V & 82.24$\pm$0.35 & 79.46$\pm$0.41 & 98.11$\pm$0.12 & 3.04$\pm$0.17 & 0.59$\pm$0.04 \\
& \textbf{VFGS-Net} & \textcolor{red}{83.23$\pm$0.33} & \textcolor{red}{82.37$\pm$0.39} & 97.87$\pm$0.13 & \textcolor{red}{2.41$\pm$0.14} & \textcolor{red}{0.55$\pm$0.03} \\
\hline
\multirow{8}{*}{HRF}
& U & 80.79$\pm$0.46 & 77.13$\pm$0.52 & 97.05$\pm$0.18 & 4.89$\pm$0.24 & 0.72$\pm$0.06 \\
& U+D & 84.52$\pm$0.38 & 83.32$\pm$0.44 & 97.02$\pm$0.16 & 2.65$\pm$0.17 & 0.47$\pm$0.04 \\
& U+B & 83.36$\pm$0.41 & 79.34$\pm$0.49 & \textcolor{red}{97.64$\pm$0.15} & 3.57$\pm$0.21 & 0.55$\pm$0.05 \\
& U+V & 84.64$\pm$0.37 & 82.59$\pm$0.43 & 97.30$\pm$0.16 & 2.71$\pm$0.18 & 0.47$\pm$0.04 \\
& U+D+B & 84.19$\pm$0.39 & 81.95$\pm$0.46 & 97.27$\pm$0.16 & 2.95$\pm$0.19 & 0.49$\pm$0.04 \\
& U+D+V & 84.87$\pm$0.36 & 83.04$\pm$0.42 & 97.27$\pm$0.15 & 2.50$\pm$0.16 & 0.45$\pm$0.03 \\
& U+B+V & 84.22$\pm$0.38 & 81.66$\pm$0.45 & 97.36$\pm$0.15 & 2.93$\pm$0.18 & 0.49$\pm$0.04 \\
& \textbf{VFGS-Net} & \textcolor{red}{85.60$\pm$0.34} & \textcolor{red}{84.35$\pm$0.40} & 97.26$\pm$0.14 & \textcolor{red}{2.21$\pm$0.15} & \textcolor{red}{0.43$\pm$0.03} \\
\hline
\multirow{8}{*}{CHASE\_DB1}
& U & 79.05$\pm$0.52 & 78.94$\pm$0.56 & 96.61$\pm$0.21 & 4.58$\pm$0.27 & 0.92$\pm$0.08 \\
& U+D & 80.13$\pm$0.48 & 79.99$\pm$0.53 & 96.80$\pm$0.19 & 3.64$\pm$0.22 & 0.81$\pm$0.07 \\
& U+B & 79.87$\pm$0.50 & 77.84$\pm$0.58 & \textcolor{red}{97.22$\pm$0.18} & 4.03$\pm$0.25 & 0.82$\pm$0.07 \\
& U+V & 80.15$\pm$0.49 & 78.67$\pm$0.55 & 97.13$\pm$0.18 & 3.88$\pm$0.23 & 0.83$\pm$0.07 \\
& U+D+B & 80.68$\pm$0.46 & 80.01$\pm$0.51 & 97.01$\pm$0.17 & 3.37$\pm$0.21 & 0.74$\pm$0.06 \\
& U+D+V & 80.02$\pm$0.47 & 78.82$\pm$0.54 & 97.03$\pm$0.18 & 3.93$\pm$0.24 & 0.82$\pm$0.07 \\
& U+B+V & 80.43$\pm$0.45 & 80.10$\pm$0.50 & 96.88$\pm$0.19 & \textcolor{red}{3.18$\pm$0.20} & 0.75$\pm$0.06 \\
& \textbf{VFGS-Net} & \textcolor{red}{81.43$\pm$0.43} & \textcolor{red}{80.69$\pm$0.48} & 97.14$\pm$0.17 & 3.60$\pm$0.22 & \textcolor{red}{0.73$\pm$0.06} \\
\hline
\multirow{8}{*}{STARE}
& U & 78.82$\pm$0.61 & 72.27$\pm$0.68 & 98.74$\pm$0.19 & 9.73$\pm$0.58 & 1.32$\pm$0.11 \\
& U+D & 78.83$\pm$0.59 & 73.00$\pm$0.66 & 98.65$\pm$0.20 & 9.95$\pm$0.61 & 1.46$\pm$0.12 \\
& U+B & 78.98$\pm$0.60 & 72.53$\pm$0.67 & \textcolor{red}{98.78$\pm$0.18} & 10.60$\pm$0.64 & 1.36$\pm$0.11 \\
& U+V & 80.10$\pm$0.57 & 76.39$\pm$0.63 & 98.48$\pm$0.21 & 6.52$\pm$0.44 & 1.04$\pm$0.09 \\
& U+D+B & 80.20$\pm$0.55 & 77.21$\pm$0.61 & 98.38$\pm$0.22 & 7.11$\pm$0.47 & 1.10$\pm$0.10 \\
& U+D+V & 80.41$\pm$0.54 & 75.57$\pm$0.62 & 98.66$\pm$0.20 & 7.49$\pm$0.49 & 1.11$\pm$0.10 \\
& U+B+V & 79.39$\pm$0.58 & 74.66$\pm$0.65 & 98.59$\pm$0.21 & 8.75$\pm$0.53 & 1.25$\pm$0.11 \\
& \textbf{VFGS-Net} & \textcolor{red}{83.21$\pm$0.36} & \textcolor{red}{82.42$\pm$0.40} & 98.39$\pm$0.14 & \textcolor{red}{3.44$\pm$0.26} & \textcolor{red}{0.76$\pm$0.07} \\
\hline
\end{tabular}
}
\end{table}

\textbf{Quantitative Results:} Table~\ref{tab:ablation} summarizes the ablation results on four public datasets, clearly illustrating the individual and complementary effects of each module. On the DRIVE dataset, introducing DFC alone improves Dice from 81.06\% to 82.40\% and reduces ASSD from 0.61 to 0.57 pixels, indicating enhanced local feature representation and boundary precision. BA-Mamba2 and VFCA individually yield limited Dice gains but contribute to improved boundary stability. Their integration leads to consistent improvements, with VFGS-Net achieving the best overall performance (Dice 83.23\%, HD95 2.41 pixels, ASSD 0.55). On the high-resolution HRF dataset, both DFC and VFCA provide substantial gains by enhancing multi-scale vessel representation and low-contrast vessel responses, while BA-Mamba2 improves topological consistency. The full VFGS-Net achieves the highest Dice (85.60\%) and lowest ASSD (0.43), demonstrating robust performance under complex vascular structures. For CHASE\_DB1 dataset, which contains fewer and noisier annotations, DFC and VFCA individually improve Dice to around 80\%, while BA-Mamba2 stabilizes boundary predictions. The complete model further improves robustness, achieving a Dice of 81.43\% and HD95 of 3.60 pixels. On STARE dataset, VFCA notably enhances the detection of fine and low-contrast vessels, DFC improves overall vessel completeness, and BA-Mamba2 strengthens global topological coherence. Their joint integration enables VFGS-Net to reach a Dice of 83.21\% and HD95 of 3.44 pixels, outperforming all ablation variants.

Overall, the ablation study confirms the complementary roles of the three components: DFC enhances local and contextual features, BA-Mamba2 captures long-range spatial dependencies, and VFCA strengthens vessel-aware frequency representations. Their integration allows VFGS-Net to consistently achieve superior performance across all datasets.

\begin{figure}[ht]
\centering
\includegraphics[width=1.00\linewidth]{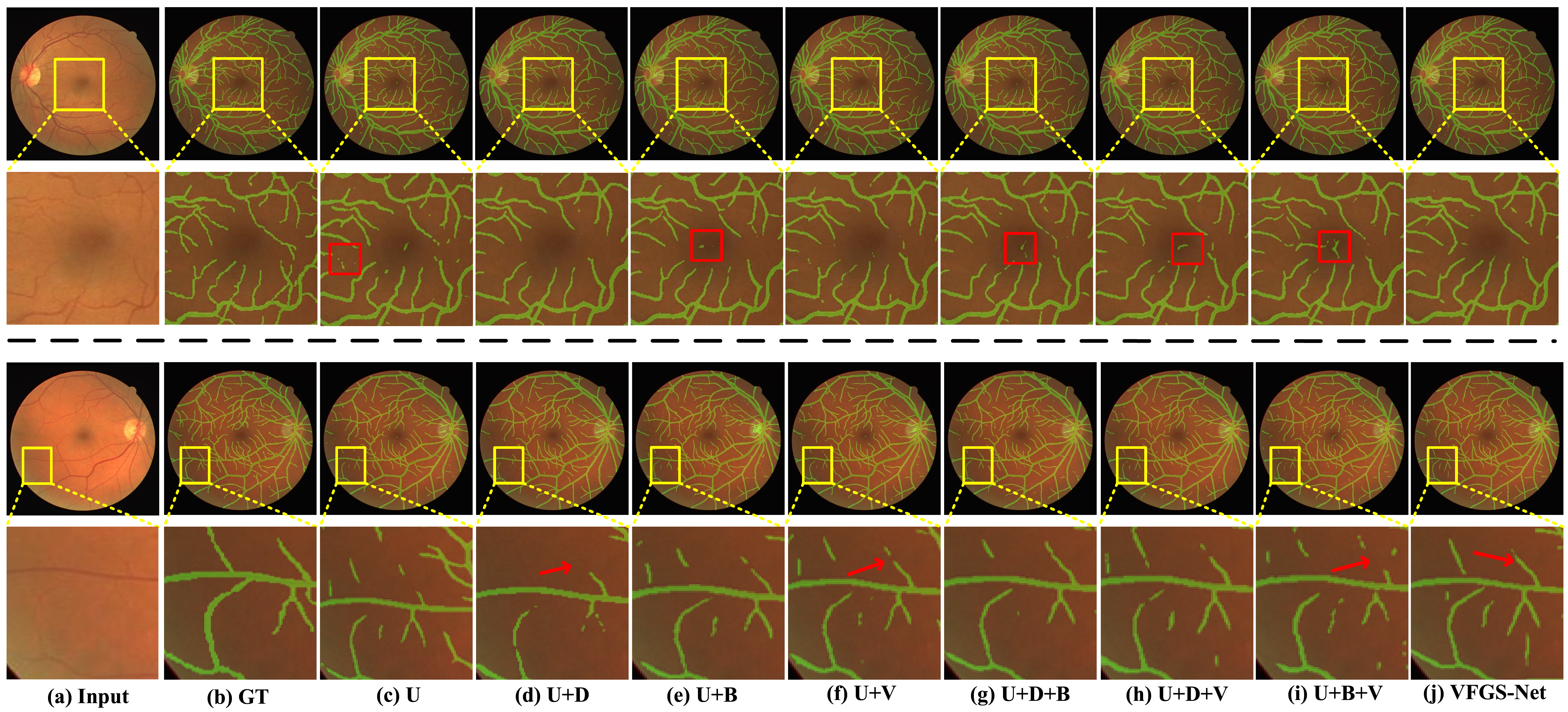}
\vspace{-2mm} 
\caption{Qualitative ablation study on the DRIVE dataset. (a-b) Input images and ground truth; (c-j) Results of different ablation variants and the complete VFGS-Net. Zoomed-in regions highlight detailed differences.}
\vspace{-2mm} 
\label{fig:ablation}
\end{figure}

\textbf{Qualitative Results:} Figure~\ref{fig:ablation} presents qualitative comparisons of different ablation variants on the DRIVE dataset. The complete VFGS-Net (Figure~\ref{fig:ablation}(j)) produces smoother vessel boundaries and more consistent vascular topology, particularly in thin branches, bifurcation regions, and highly tortuous vessels, while effectively suppressing spurious artifacts. In contrast, variants with only one or two modules (Figures~\ref{fig:ablation}(c-i)) tend to suffer from vessel discontinuities, false positives, or blurred boundaries in challenging regions, as highlighted by the red rectangles. The DFC (D) module enhances local texture modeling and contextual feature aggregation, leading to more coherent representations of major vessels. The BA-Mamba2 (B) module strengthens long-range dependency modeling and alleviates vessel breakage at bifurcations, while the VFCA (V) module improves the discriminability of microvessels and low-contrast branches. When all three modules are jointly integrated, VFGS-Net is able to preserve both large vessels and fine capillaries simultaneously. Regions marked by red arrows further demonstrate its superior capability in accurately delineating tortuous and tiny vessels, clearly outperforming other ablation variants.

\section{Discussions}

\subsection{Segmentation Performance}
Quantitative results on four public retinal vessel datasets (Table~\ref{tab:quantitative}) show that VFGS-Net consistently outperforms six state-of-the-art methods in Dice, HD95, and ASSD, demonstrating its effectiveness in capturing fine vessels while preserving global vascular continuity. The performance gains across datasets with diverse imaging conditions indicate strong robustness to vessel width variation, complex branching, and pathological interference.

\begin{figure}[ht]
\centering
\includegraphics[width=1.00\linewidth]{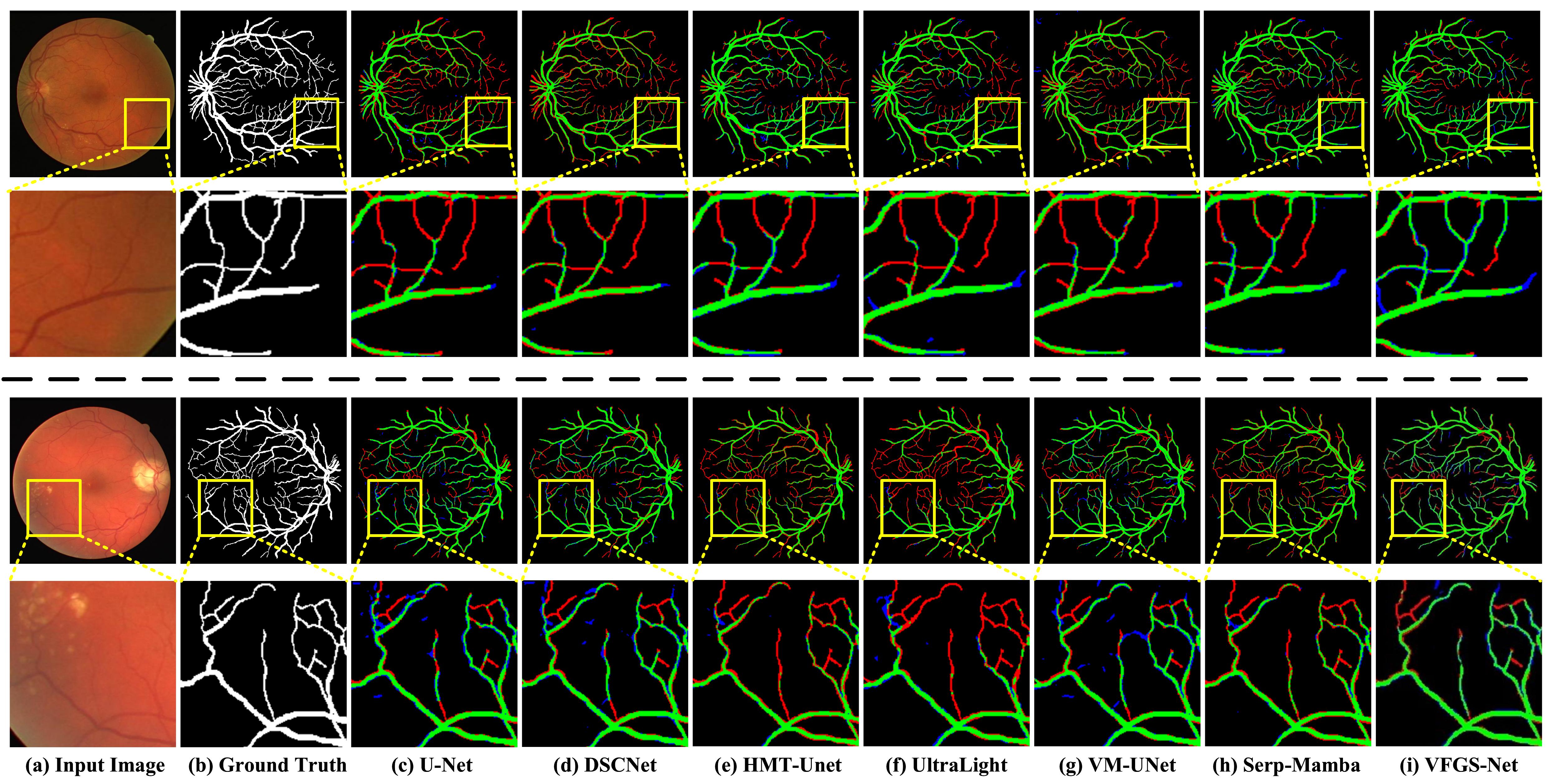}
\vspace{-2mm} 
\caption{Difference map comparison among retinal vessel segmentation methods. (a) Input images; (b) Ground truth; (c–i) Difference maps produced by different models.
True positives, false negatives, and false positives are shown in green, red, and blue, respectively.}
\vspace{-2mm} 
\label{fig:diffmap}
\end{figure}

Qualitative difference maps in Figure~\ref{fig:diffmap} further confirm these results. Compared with competing methods, VFGS-Net produces fewer false negatives and false positives, particularly around major vessels, bifurcations, and highly curved regions, reflecting improved continuity and boundary accuracy.

These advantages stem from the complementary design of DFC, BA-Mamba2, and VFCA. DFC strengthens local feature representation, BA-Mamba2 models long-range vascular topology, and VFCA enhances the response of microvessels and low-contrast branches. Together, they effectively reduce vessel fragmentation and segmentation artifacts, enabling accurate delineation across both major vessels and distal branches.

\subsection{Clinical Implications}
Accurate retinal vessel segmentation is crucial for the diagnosis and monitoring of microvascular diseases such as diabetic and hypertensive retinopathy, yet remains challenging due to thin vessels, complex bifurcations, and low-contrast regions.

By jointly modeling local details, global topology, and frequency-domain vessel characteristics, VFGS-Net provides a reliable automated solution for retinal vessel delineation. Its robust segmentation performance enables more accurate extraction of clinically relevant biomarkers, including vessel density, branching patterns, and diameter-related measurements, supporting early screening, disease assessment, and treatment planning.

\section{Conclusions}

This work presents VFGS-Net, a novel network for retinal vessel segmentation that jointly models local texture information, long-range vascular topology, and vessel-aware frequency characteristics. By integrating DFC, BA-Mamba2, and VFCA, the proposed method effectively addresses key challenges such as low contrast, noise interference, and the accurate delineation of fine vessels. Extensive experiments on multiple public datasets demonstrate that VFGS-Net consistently outperforms representative state-of-the-art methods in Dice, HD95, and ASSD, achieving improved vessel continuity, boundary accuracy, and preservation of distal branches. Both quantitative and qualitative results confirm its robustness across diverse imaging conditions. Overall, VFGS-Net provides a reliable and effective solution for retinal vessel segmentation, with promising potential for supporting retinal disease screening, vascular morphology analysis, and image-assisted clinical diagnosis.

\section*{Acknowledgments}
This work was supported by the National Natural
Science Foundation of China (62006165) and the Natural Science Foundation of Sichuan Province (2025ZNSFSC1477).

\bibliographystyle{unsrt}  


\begin{thebibliography}{99}

\bibitem{yau2012global}
Joanne~W.~Y. Yau, Sophie~L. Rogers, Ryo Kawasaki, Ecosse~L. Lamoureux, Jonathan~W. Kowalski, Toke Bek, Shih-Jen Chen, Jacqueline~M. Dekker, Astrid Fletcher, Jakob Grauslund, \emph{et al.}
\newblock Global prevalence and major risk factors of diabetic retinopathy.
\newblock {\em Diabetes Care}, 35(3):556--564, 2012.

\bibitem{wong2007eye}
Tien Wong and Paul Mitchell.
\newblock The eye in hypertension.
\newblock {\em The Lancet}, 369(9559):425--435, 2007.

\bibitem{flammer1994vascular}
Josef Flammer.
\newblock The vascular concept of glaucoma.
\newblock {\em Survey of Ophthalmology}, 38:S3--S6, 1994.

\bibitem{cheung2012retinal}
Carol~Yim-lui Cheung, M.~Kamran Ikram, Charumathi Sabanayagam, and Tien~Yin Wong.
\newblock Retinal microvasculature as a model to study the manifestations of hypertension.
\newblock {\em Hypertension}, 60(5):1094--1103, 2012.

\bibitem{fraz2012blood}
Muhammad~Moazam Fraz, Paolo Remagnino, Andreas Hoppe, Bunyarit Uyyanonvara, Alicja~R. Rudnicka, Christopher~G. Owen, and Sarah~A. Barman.
\newblock Blood vessel segmentation methodologies in retinal images---a survey.
\newblock {\em Computer Methods and Programs in Biomedicine}, 108(1):407--433, 2012.

\bibitem{verma2024systematic}
Prem~Kumari Verma and Jagdeep Kaur.
\newblock Systematic review of retinal blood vessels segmentation based on AI-driven technique.
\newblock {\em Journal of Imaging Informatics in Medicine}, 37(4):1783--1799, 2024.

\bibitem{qin2024review}
Qing Qin and Yuanyuan Chen.
\newblock A review of retinal vessel segmentation for fundus image analysis.
\newblock {\em Engineering Applications of Artificial Intelligence}, 128:107454, 2024.

\bibitem{lecun2002gradient}
Yann LeCun, L{\'e}on Bottou, Yoshua Bengio, and Patrick Haffner.
\newblock Gradient-based learning applied to document recognition.
\newblock {\em Proceedings of the IEEE}, 86(11):2278--2324, 2002.

\bibitem{ronneberger2015u}
Olaf Ronneberger, Philipp Fischer, and Thomas Brox.
\newblock U-net: Convolutional networks for biomedical image segmentation.
\newblock In {\em International Conference on Medical Image Computing and Computer-Assisted Intervention}, pages 234--241. Springer, 2015.

\bibitem{chen2021transunet}
Jieneng Chen, Yongyi Lu, Qihang Yu, Xiangde Luo, Ehsan Adeli, Yan Wang, Le Lu, Alan~L. Yuille, and Yuyin Zhou.
\newblock Transunet: Transformers make strong encoders for medical image segmentation.
\newblock {\em arXiv preprint arXiv:2102.04306}, 2021.

\bibitem{ruan2024vm}
Jiacheng Ruan, Jincheng Li, and Suncheng Xiang.
\newblock Vm-unet: Vision mamba unet for medical image segmentation.
\newblock {\em ACM Transactions on Multimedia Computing, Communications and Applications}, 2024.

\bibitem{cao2022swin}
Hu Cao, Yueyue Wang, Joy Chen, Dongsheng Jiang, Xiaopeng Zhang, Qi Tian, and Manning Wang.
\newblock Swin-unet: Unet-like pure transformer for medical image segmentation.
\newblock In {\em European Conference on Computer Vision}, pages 205--218. Springer, 2022.

\bibitem{mu2023attention}
Nan Mu, Zonghan Lyu, Mostafa Rezaeitaleshmahalleh, Jinshan Tang, and Jingfeng Jiang.
\newblock An attention residual u-net with differential preprocessing and geometric postprocessing: Learning how to segment vasculature including intracranial aneurysms.
\newblock {\em Medical Image Analysis}, 84:102697, 2023.

\bibitem{mu2023exploring}
Nan Mu, Zonghan Lyu, Xiaoming Zhang, Robert McBane, Aditya~S. Pandey, and Jingfeng Jiang.
\newblock Exploring a frequency-domain attention-guided cascade U-Net: Towards spatially tunable segmentation of vasculature.
\newblock {\em Computers in Biology and Medicine}, 167:107648, 2023.

\bibitem{gu2021efficiently}
Albert Gu, Karan Goel, and Christopher R{\'e}.
\newblock Efficiently modeling long sequences with structured state spaces.
\newblock {\em arXiv preprint arXiv:2111.00396}, 2021.

\bibitem{gu2024mamba}
Albert Gu and Tri Dao.
\newblock Mamba: Linear-time sequence modeling with selective state spaces.
\newblock In {\em First Conference on Language Modeling}, 2024.

\bibitem{wang2024mamba}
Ziyang Wang, Jian-Qing Zheng, Yichi Zhang, Ge Cui, and Lei Li.
\newblock Mamba-unet: Unet-like pure visual mamba for medical image segmentation.
\newblock {\em arXiv preprint arXiv:2402.05079}, 2024.

\bibitem{wu2025ultralight}
Renkai Wu, Yinghao Liu, Guochen Ning, Pengchen Liang, and Qing Chang.
\newblock Ultralight vm-unet: Parallel vision mamba significantly reduces parameters for skin lesion segmentation.
\newblock {\em Patterns}, 6(11), 2025.

\bibitem{zhang2024hmt}
Mingya Zhang, Zhihao Chen, Yiyuan Ge, and Xianping Tao.
\newblock HMT-UNet: A hybrid mamba-transformer vision U-Net for medical image segmentation.
\newblock {\em arXiv preprint arXiv:2408.11289}, 2024.

\bibitem{li2020iternet}
Liangzhi Li, Manisha Verma, Yuta Nakashima, Hajime Nagahara, and Ryo Kawasaki.
\newblock Iternet: Retinal image segmentation utilizing structural redundancy in vessel networks.
\newblock In {\em Proceedings of the IEEE/CVF Winter Conference on Applications of Computer Vision}, pages 3656--3665, 2020.

\bibitem{mou2021cs2}
Lei Mou, Yitian Zhao, Huazhu Fu, Yonghuai Liu, Jun Cheng, Yalin Zheng, Pan Su, Jianlong Yang, Li Chen, Alejandro~F. Frangi, \emph{et al.}
\newblock CS$^2$-Net: Deep learning segmentation of curvilinear structures in medical imaging.
\newblock {\em Medical Image Analysis}, 67:101874, 2021.

\bibitem{qi2023dynamic}
Yaolei Qi, Yuting He, Xiaoming Qi, Yuan Zhang, and Guanyu Yang.
\newblock Dynamic snake convolution based on topological geometric constraints for tubular structure segmentation.
\newblock In {\em Proceedings of the IEEE/CVF International Conference on Computer Vision}, pages 6070--6079, 2023.

\bibitem{wang2025serp}
Hongqiu Wang, Yixian Chen, Wu Chen, Huihui Xu, Haoyu Zhao, Bin Sheng, Huazhu Fu, Guang Yang, and Lei Zhu.
\newblock Serp-mamba: Advancing high-resolution retinal vessel segmentation with selective state-space model.
\newblock {\em IEEE Transactions on Medical Imaging}, 2025.

\bibitem{dao2024transformers}
Tri Dao and Albert Gu.
\newblock Transformers are SSMs: Generalized models and efficient algorithms through structured state space duality.
\newblock {\em arXiv preprint arXiv:2405.21060}, 2024.

\bibitem{selvaraju2017grad}
Ramprasaath~R. Selvaraju, Michael Cogswell, Abhishek Das, Ramakrishna Vedantam, Devi Parikh, and Dhruv Batra.
\newblock Grad-CAM: Visual explanations from deep networks via gradient-based localization.
\newblock In {\em Proceedings of the IEEE International Conference on Computer Vision}, pages 618--626, 2017.

\bibitem{kingma2014adam}
Diederik~P. Kingma and Jimmy Ba.
\newblock Adam: A method for stochastic optimization.
\newblock {\em arXiv preprint arXiv:1412.6980}, 2014.

\bibitem{shaout2023novel}
Adnan Shaout and Jiho Han.
\newblock A novel retinal image contrast enhancement---fuzzy-based method.
\newblock In {\em 2023 24th International Arab Conference on Information Technology (ACIT)}, pages 1--6. IEEE, 2023.

\bibitem{fraz2012ensemble}
Muhammad~Moazam Fraz, Paolo Remagnino, Andreas Hoppe, Bunyarit Uyyanonvara, Alicja~R. Rudnicka, Christopher~G. Owen, and Sarah~A. Barman.
\newblock An ensemble classification-based approach applied to retinal blood vessel segmentation.
\newblock {\em IEEE Transactions on Biomedical Engineering}, 59(9):2538--2548, 2012.

\bibitem{staal2004ridge}
Joes Staal, Michael~D. Abr{\`a}moff, Meindert Niemeijer, Max~A. Viergever, and Bram van Ginneken.
\newblock Ridge-based vessel segmentation in color images of the retina.
\newblock {\em IEEE Transactions on Medical Imaging}, 23(4):501--509, 2004.

\bibitem{budai2013robust}
Attila Budai, R{\"u}diger Bock, Andreas Maier, Joachim Hornegger, and Georg Michelson.
\newblock Robust vessel segmentation in fundus images.
\newblock {\em International Journal of Biomedical Imaging}, 2013:154860, 2013.

\bibitem{hoover2000locating}
A.~D. Hoover, Valentina Kouznetsova, and Michael Goldbaum.
\newblock Locating blood vessels in retinal images by piecewise threshold probing of a matched filter response.
\newblock {\em IEEE Transactions on Medical Imaging}, 19(3):203--210, 2000.

\end{thebibliography}

\end{document}